\pdfoutput=1

\documentclass[11pt]{article}

\usepackage{emnlp2021}

\usepackage{latexsym}
\usepackage[T1]{fontenc}

\usepackage[utf8]{inputenc}
\usepackage{times}
\usepackage{soul}
\usepackage{url}
\usepackage{graphicx}
\usepackage{amsmath}
\usepackage{amsthm}
\usepackage{booktabs}
\usepackage{algorithm}
\usepackage{algorithmicx}
\usepackage{algpseudocode}
\usepackage{microtype}
\usepackage{hyperref}
\usepackage{caption}
\usepackage{arydshln}
\usepackage{pgfplots} 
\usepackage{verbatim}
\floatname{algorithm}{Algorithm}

\title{Improving Graph-based Sentence Ordering with \\Iteratively Predicted Pairwise Orderings}

\author{
Shaopeng Lai$^1$, Ante Wang$^1$, Fandong Meng$^2$, Jie Zhou$^2$, Yubin Ge$^3$, Jiali Zeng$^4$, \\ \bf{Junfeng Yao}$^1$, Degen Huang$^5$ \and \bf{Jinsong Su}$^{1,6 }$\thanks{\quad Corresponding author} \\
$^{1}$Xiamen University, China  $^{2}$Pattern Recognition Center, WeChat AI, Tencent Inc, China \\ $^{3}$University of Illinois at Urbana-Champaign, USA $^{4}$Tencent Cloud Xiaowei,  China \\ $^{5}$Dalian University of Technology, China $^{6}$Pengcheng  Laboratory,  China\\
splai@stu.xmu.edu.cn, 
fandongmeng@tencent.com, 
jssu@xmu.edu.cn
}

\begin{document}

\maketitle

\begin{abstract}
	Dominant sentence ordering models can be classified into pairwise ordering models and set-to-sequence models. However, there is little attempt to combine these two types of models, which inituitively possess complementary advantages. In this paper, we propose a novel sentence ordering framework which introduces two classifiers to make better use of pairwise orderings for graph-based sentence ordering \cite{Yin:IJCAI19,YinLSP_JAIR20}. Specially, given an initial sentence-entity graph, we first introduce a graph-based classifier to predict pairwise orderings between linked sentences. Then, in an iterative manner, based on the graph updated by previously predicted high-confident pairwise orderings, another classifier is used to predict the remaining uncertain pairwise orderings. At last, we adapt a GRN-based sentence ordering model \cite{Yin:IJCAI19,YinLSP_JAIR20} on the basis of final graph. Experiments on five commonly-used datasets demonstrate the effectiveness and generality of our model. Particularly, when equipped with BERT \cite{Devlin:BERT} and FHDecoder \cite{Yin:AAAI20}, our model achieves state-of-the-art performance. Our code is available at \url{https://github.com/DeepLearnXMU/IRSEG}.
\end{abstract}

\section{Introduction}

With the rapid development and increasing applications of natural language processing (NLP), modeling text coherence has become a significant task, since it can provide beneficial information for understanding, evaluating and generating multi-sentence texts. As an important subtask, sentence ordering aims at recovering unordered sentences back to naturally coherent paragraphs. It is required to deal with logic and syntactic consistency, and has increasingly attracted attention due to its wide applications on several tasks such as text generation \cite{Konstas:NAACL12,Holtzman:ACL2018} and extractive summarization \cite{Barzilay:jair02,Nallapati:aaai12}.

Recently, inspired by the great success of deep learning in other NLP tasks, researchers have resorted to neural sentence ordering models, which can be classified into: \emph{pairwise ordering models} \cite{Chen16,Agrawal:ACL16,Li:emnlp17,Moon:EMNLP19,Kumar:aaai20,Prabhumoye:ACL20,Zhu:AAAI2021} and \emph{set-to-sequence models} \cite{Gong16,Nguyen:acl17,Logeswaran:aaai18,Mohiuddin:acl18,Cui:emnlp18,Yin:IJCAI19,Oh:emnlp19,Yin:AAAI20,Cui:emnlp20,YinLSP_JAIR20}. Generally, the former predicts the relative orderings between pairwise sentences, which are then leveraged to produce the final ordered sentence sequence. Its advantage lies in the lightweight pairwise ordering predictions, since the predictions only depend on the semantic representations of involved sentences. By contrast, the latter is mainly based on an encoder-decoder framework, where an encoder is first used to learn contexualized sentence representations by considering other sentences, and then a decoder, such as pointer network \cite{Vinyals:NIPS15}, outputs ordered sentences.

Overall, these two kinds of models have their own strengths, which are complementary to each other. To combine their advantages, \citet{Yin:AAAI20} propose FHDecoder that is equipped with three pairwise ordering prediction modules to enhance the pointer network decoder. Along this line, \citet{Cui:emnlp20} introduce BERT to exploit the deep semantic connection and relative orderings between sentences and achieve SOTA performance when equipped with FHDecoder. However, there still exist two drawbacks:
1) their pairwise ordering predictions only depend on involved sentence pairs, without considering other sentences in the same set;
2) their one-pass pairwise ordering predictions are relatively rough, ignoring distinct difficulties in predicting different sentence pairs. Therefore, we believe that the potential of pairwise orderings in neural sentence ordering models has not been fully exploited. 

In this paper, we propose a novel iterative pairwise ordering prediction framework which introduces two classifiers to make better use of pairwise orderings for graph-based sentence ordering \cite{Yin:IJCAI19,YinLSP_JAIR20}. As an extension of \emph{Sentence-Enity Graph Recurrent Network} (SE-GRN) \cite{Yin:IJCAI19,YinLSP_JAIR20}, our framework enriches the graph representation with iteratively predicted orderings between pairwise sentences, which further benefits the subsequent generation of ordered sentences. The basic intuitions behind our work are two-fold. First, learning contextual sentence representations is helpful to predict pairwise orderings. Second, difficulties of predicting ordering vary with respect to different sentence pairs. Thus, it is more reasonable to first predict the orderings of pairwise sentences easily to be predicted, and then leverage these predicted orderings to refine the predictions for other pairwise sentences. 

Concretely, we propose two graph-based classifiers to iteratively conduct ordering predictions for pairwise sentences. The first classifier takes the sentence-entity graph (SE-Graph) \cite{Yin:IJCAI19,YinLSP_JAIR20} as input and yields relative orderings of linked sentences via corresponding probabilities. Next, in an iterative manner, the second classifier enriches the previous graph representation by converting high-value probabilities into the weights of the corresponding edges, and then reconduct graph encoding to predict orderings for the other pairwise sentences. Based on the final weighted graph representation, we adapt SE-GRN to construct a graph-based sentence ordering model, of which the decoder is also a pointer network. 

To the best of our knowledge, our work is the first to exploit pairwise orderings to enhance the graph encoding for graph-based set-to-squence sentence ordering. To investigate the effectiveness of our framework, we conduct extensive experiments on several commonly-used datasets. Experimental results and in-depth analyses show that our model enhanced with some proposed technologies \cite{Devlin:BERT,Yin:AAAI20} achieves the state-of-the-art performance.

\section{Related Work}\label{section:related wrok}

Early studies mainly focused on exploring human-designed features for sentence ordering \cite{Lapata:acl03,Barzilay:naacl04,Barzilay:ACL05,Barzilay:CL,Elsner:acl11,Guinaudeau:acl13}.
Recently, neural network based sentence ordering models have become dominant , consisting of the following two kinds of models:

1) \textbf{Pairwise models}. Generally, they first predict the pairwise orderings between sentences and then use them to produce the final sentence order via ranking algorithms \cite{Chen16,Agrawal:ACL16,Li:emnlp17,Kumar:aaai20,Prabhumoye:ACL20,Zhu:AAAI2021}.
For example, \citet{Chen16} first framed sentence ordering as a ranking task conditioned on pairwise scores. \citet{Agrawal:ACL16} conducted the same experiments as \cite{Chen16} in the task of image caption storytelling. Similarly, \citet{Li:emnlp17} investigated the effectiveness of discriminative and generative models on ordering pairs of sentences in small domains. \citet{Moon:EMNLP19} proposed a unified model that incorporates sentence grammar, pairwise coherence and global coherence into a common neural framework. Recently, \citet{Prabhumoye:ACL20} and \citet{Zhu:AAAI2021} employed ranking techniques to find the right order of the sentences under the constraint of the predicted pairwise sentence ordering;

2) \textbf{Set-to-sequence Models}. Basically, these models are based on an encoder-decoder framework, where the encoder is used to obtain sentence representations and then the decoder produces ordered sentences progressively. Among them, both \citet{Gong16} and \citet{Logeswaran:aaai18} explored RNN based encoder, while both \citet{Nguyen:acl17} and \citet{Mohiuddin:acl18} employed neural entity grid models as encoders. Typically, \citet{Cui:emnlp18} proposed ATTOrderNet that uses self-attention mechanism to learn sentence representations. Inspired by the successful applications of graph neural network (GNN) in many NLP tasks \cite{song2018exploring,XueIJCAI19,Songtacl19,Song:acl20}, \citet{Yin:IJCAI19,YinLSP_JAIR20} represented input sentences with a unified SE-Graph and then applied GRN to learn sentence representations. Very recently, we notice that \citet{Somnath:arxiv} proposes a BART-based sentence ordering model. Please note that our porposed framework is compatible with BART \cite{Liu:acl20:bart}. For example, we can easily adapt the BART encoder as our sentence encoder.

With similar motivation with ours, that is, to combine advantages of above-mentioned two kinds of models, \citet{Yin:AAAI20} introduced three pairwise ordering predicting modules (FHDecoder) to enhance the pointer network decoder of ATTOrderNet. 
Recently, \citet{Cui:emnlp20} proposed BERSON that is also equipped with FHDecoder and utilizes BERT to exploit the deep semantic connection and relative ordering between sentences.

However, significantly different from them, we borrow the idea from the mask-predict framework \cite{Gu:ICLR18,Ghazvininejad:EMNLP-IJNLP19,Deng:ECCV20} to progressively incorporate pairwise ordering information into SE-Graph, which is the basis of our graph-based sentence ordering model.
To the best of our knowledge, our work is the first attempt to explore iteratively refined GNN for sentence ordering.

\section{Background}

\begin{figure}[!t]
	\includegraphics[width=1\linewidth]{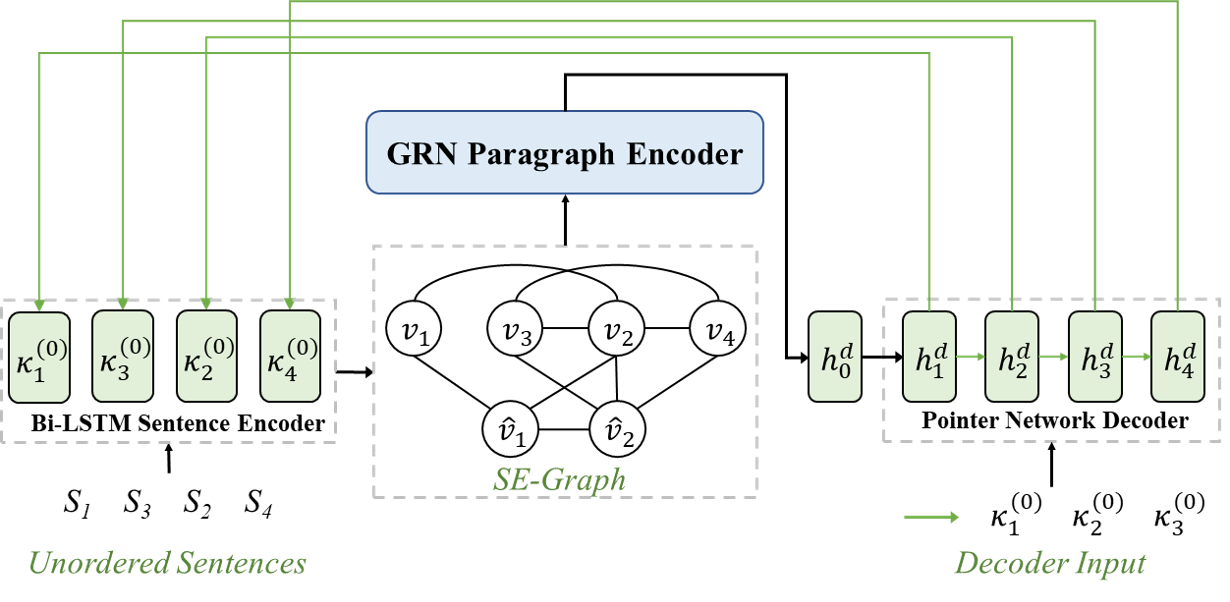}
	\caption{\label{segraph}The architecture of SE-GRN model \cite{Yin:IJCAI19,YinLSP_JAIR20}.}

\end{figure}

In this section, we give a brief introduction to the SE-GRN \cite{Yin:IJCAI19,YinLSP_JAIR20}, which is selected as our baseline due to its competitive performance. As shown in Figure \ref{segraph}, SE-GRN is composed of a Bi-LSTM sentence encoder, GRN \cite{Zhang:acl18} paragraph encoder, and a pointer network \cite{Vinyals} decoder.

\subsection{Sentence-Entity Graph}\label{subsection segraph}
The SE-GRN takes $I$ sentences $\boldsymbol{s}=\left[s_{o_1},\dots,s_{o_I}\right]$ as input and tries to predict their correct order $\boldsymbol{o}^*=\left[o_1^*,\dots,o_I^*\right]$. At first, each sentence $s_{o_i}$ is fed into a Bi-LSTM sentence encoder, where the concatenation of the last hidden states in two directions is used as the context-aware sentence representation $\boldsymbol{\kappa}^\text{(0)}_{o_i}$.
As illustrated in the middle of Figure \ref{segraph}, each input sentence set is represented as an undirected sentence-entity graph $G=(\boldsymbol{V},\boldsymbol{E})$, where $\boldsymbol{V}$$=$$\{v_i\}_{i=1}^I$$\cup$$\{\hat{v}_j\}_{j=1}^J$ and $\boldsymbol{E}=$$\{e_{i,i'}\}_{i=1,i'=1}^{I,I}\cup\{\bar{e}_{i,j}\}_{i=1,j=1}^{I,J}\cup\{\hat{e}_{j,j'}\}_{j=1,j'=1}^{J,J}$ represent the nodes and edges respectively. Here, nodes include \emph{sentence nodes} (such as $v_i$) and \emph{entity nodes} (such as $\hat{v}_j$), and each edge is
1) \emph{sentence-sentence edge} (\textit{ss-edge}, such as $e_{i,i'}$) linking two sentences having the same entity; 
or 2) \emph{sentence-entity edge} (\textit{se-edge}, such as $\bar{e}_{i,j}$) connecting an entity to a sentence that contains it. 
Each se-edge is assigned with a label including \emph{subject}, \emph{object} or \emph{other}, based on the syntactic role of its involved entity; 
or 3) \emph{entity-entity edge} (\emph{ee-edge}, such as $\hat{e}_{j,j'}$) connecting two semantic related entities.
Besides, a virtual global node connecting to all nodes is introduced to capture global information effectively.

\begin{figure*}[t]
	\centering
	\includegraphics[width=1\linewidth]{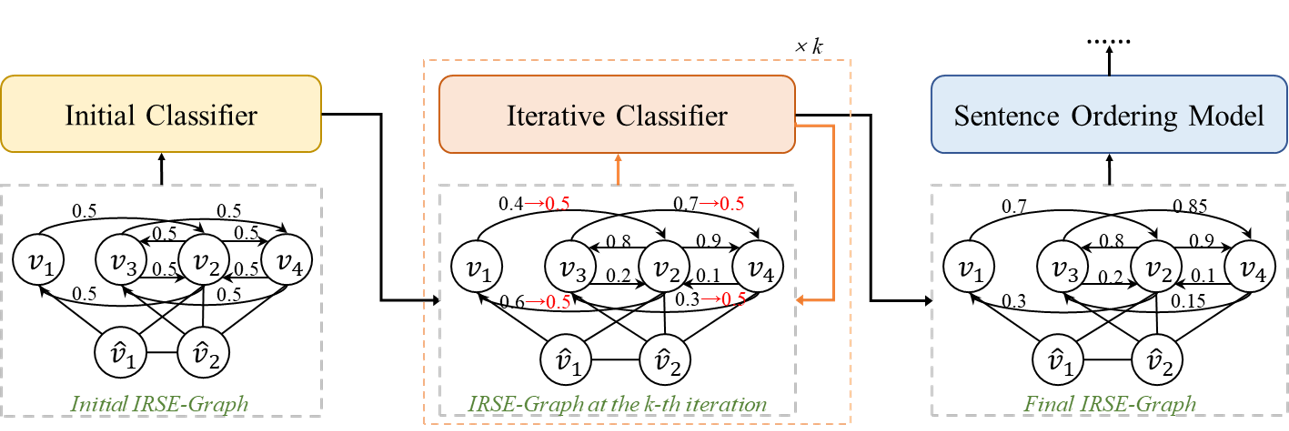}
	\caption{\label{pipeline} The architecture of our model during inference. IRSE-Graph is a weighted graph representation, of which weights of ss-edges are iteratively refined by iterative classifier. Note that we construct the sentence ordering model based on the final IRSE-Graph.}

\end{figure*}
\subsection{Paragraph Encoding with GRN} \label{subsection grn encoding}
Node representations of each sentence and each entity are first initialized with the concatenation of bidirectional last states of the Bi-LSTM sentence encoder and the corresponding GloVe word embedding, respectively.
Then, a GRN is adapted to encode the above sentence-entity graph, where node states are updated iteratively.
During the process of updating hidden states, the messages for each node are aggregated from its adjacent nodes.
Specifically, the sentence-level message $\boldsymbol{m}_i^\text{($l$)}$ and entity-level message
$\boldsymbol{\tilde{m}}_i^\text{($l$)}$ for a sentence $s_i$ are defined as follows:

\begin{small}
	\begin{equation} \label{eq:egraph_s_mess}
		\begin{aligned}
			\boldsymbol{m}_i^\text{($l$)}&=\sum_{v_{i^\prime} \in N_i} w(\boldsymbol{\kappa}_{i}^\text{($l$-1)},\boldsymbol{\kappa}_{i^\prime}^\text{($l$-1)})\boldsymbol{\kappa}_{i^\prime}^{\text{($l$-1)}}, \\ 
			\boldsymbol{\hat{m}}_i^\text{($l$)}&=\sum_{{v}_j \in \hat{N}_i} \bar{w}(\boldsymbol{\kappa}_{i}^\text{($l$-1)},\boldsymbol{\epsilon}_{j}^\text{($l$-1)},\boldsymbol{r}_{ij})\boldsymbol{\epsilon}_j^\text{($l$-1)},
		\end{aligned}
	\end{equation}
\end{small}where $\boldsymbol{\kappa}_{i^\prime}^\text{($l$-1)}$ and $\boldsymbol{\epsilon}_j^\text{($l$-1)}$ stand for the neighboring sentence and entity representations of the $i$-th sentence node $v_i$ at the $(l-1)$-th layer, $N_i$ and $\hat{N}_i$ denote the sets of neighboring sentences and entities of $v_i$, and both $w(*)$ and $\bar{w}(*)$ are gating functions with single-layer networks, involving associated node states and edge label $r_{ij}$ (if any).

Afterwards, $\boldsymbol{\kappa}_i^\text{($l$-1)}$ is updated by concatenating its original representation $\boldsymbol{\kappa}_{i}^\text{(0)}$, the messages from neighbours ($\boldsymbol{m}_i^\text{($l$)}$ and$\  \boldsymbol{\hat{m}}_i^\text{($l$)}$) and the global state $\boldsymbol{g}^\text{($l$-1)}$ via GRU:
\begin{small}
	\begin{equation} \label{eq:egraph_s_gate}
		\begin{aligned}
			& \boldsymbol{\xi}_i^\text{($l$)}=[\boldsymbol{\kappa}_{i}^\text{(0)};\boldsymbol{m}_i^\text{($l$)};\boldsymbol{\hat{m}}_i^\text{($l$)};\boldsymbol{g}^\text{($l$-1)}],\\
			& \boldsymbol{\kappa}_i^\text{($l$)}=\text{GRU}(\boldsymbol{\xi}_i^\text{($l$)}, \boldsymbol{\kappa}_{i^\prime}^\text{($l$-1)}).
		\end{aligned}
	\end{equation}
\end{small}Similar to updating sentence nodes, each entity state $\boldsymbol{\epsilon}_j^\text{($l$-1)}$ is updated based on its word embedding $\boldsymbol{emb}_j$, hidden states of its connected sentence nodes (such as $\boldsymbol{\kappa}_i^\text{($l$-1)}$), and $\boldsymbol{g}^\text{($l$-1)}$:
\begin{small}
	\begin{equation} \label{eq:egraph_e_gate}
		\begin{aligned}
			\boldsymbol{m}_j^\text{($l$)}&=\sum_{v_i \in {N}_j} \bar{w}(\boldsymbol{\epsilon}_{j}^\text{($l$-1)},\boldsymbol{\kappa}_{i}^\text{($l$-1)},\boldsymbol{r}_{ij}) \boldsymbol{\kappa}_i^\text{($l$-1)},\\
			\boldsymbol{\hat{m}}_j^\text{($l$)}&=\sum_{{v}_{j'} \in \hat{N}_j} \tilde{w}(\boldsymbol{\epsilon}_{j}^\text{($l$-1)},\boldsymbol{\epsilon}_{j'}^\text{($l$-1)})\boldsymbol{\epsilon}_j^\text{($l$-1)},\\
			\boldsymbol{{\xi}}_j^\text{($l$)}&=[\boldsymbol{emb}_j;\boldsymbol{{m}}_j^\text{($l$)};\boldsymbol{\hat{m}}_j^\text{($l$)};\boldsymbol{g}^\text{($l$-1)}],\\
			\boldsymbol{\epsilon}_j^\text{($l$)}&=\text{GRU}(\boldsymbol{{\xi}}_j^\text{($l$)}, \boldsymbol{\epsilon}_j^\text{($l$-1)}).
		\end{aligned}
	\end{equation}
\end{small}Finally, the messages from both sentence and entity states are used to update global state $\boldsymbol{g}^\text{($l$-1)}$ via
\begin{small}
	\begin{equation} \label{eq:egraph_e_gate}
			\boldsymbol{g}^\text{($l$)}=
			\text{GRU}(\frac{1}{|\boldsymbol{V}|}\sum_{v_i \in \boldsymbol{V}} \boldsymbol{\kappa}_i^\text{($l$-1)}, 
			\frac{1}{|\boldsymbol{\hat{V}}|}\sum_{\hat{v}_j \in \boldsymbol{\hat{V}}} \boldsymbol{\epsilon}_j^\text{($l$-1)}, 
			\boldsymbol{g}^\text{($l$-1)}).
	\end{equation}
\end{small}

The above updating process is iterated for $L$ times. Usually, the top hidden states are considered as fine-grained graph representations, which will provide dynamical context for the decoder via attention mechanism. 

\subsection{Decoding with Pointer Network}
Given the learned hidden states $\{\boldsymbol{\kappa}^\text{($L$)}_i\}$ and $\boldsymbol{g}^\text{($L$)}$, the prediction procedure for order $\boldsymbol{o}'$ can be formalized as follows:
\begin{small}
	\begin{equation}
		\begin{aligned}
			P(\boldsymbol{o}'|\boldsymbol{K}^\text{($L$)}) &= \prod_{t=1}^{\boldsymbol{I}} P(o'_t|\boldsymbol{o}'_{<t},\boldsymbol{K}^\text{($L$)}_{o'_{t-1}}), \\
			P(o'_t|\boldsymbol{o}'_{<t},\boldsymbol{K}^\text{($L$)}_{o'_{t-1}})&=\textrm{softmax}(\boldsymbol{q}^\mathrm{T}\tanh(\boldsymbol{W}\boldsymbol{h}_t^{d} + \boldsymbol{U}\boldsymbol{K}^\text{($L$)}_{o'_{t-1}})), \notag\\
		\end{aligned}
	\end{equation}
	\begin{equation}
		\begin{aligned}
			\boldsymbol{h}_t^{d} &= \textrm{LSTM}(\boldsymbol{h}_{t-1}^{d}, \boldsymbol{\kappa}^\text{(0)}_{o'_{t-1}}).
		\end{aligned}
	\end{equation}
\end{small}Here, $\boldsymbol{q}$, $\boldsymbol{W}$ and $\boldsymbol{U}$ are learnable parameters,  $\boldsymbol{K}^\text{($L$)}_{o'_{t-1}}$ and $\boldsymbol{h}_t^{d}$ denote the sentence representations with predicted order  $\left[\boldsymbol{\kappa}^\text{($L$)}_{o_1'},\dots,\boldsymbol{\kappa}^\text{($L$)}_{o_{t-1}'}\right]$ and the decoder hidden state at the $t$-th time step, which is initialized by $\boldsymbol{g}^\text{($L$)}$ as $t$$=$$0$, respectively.

\section{Our Framework}
In this section, we give a detailed description to our framework.
As shown in Figure \ref{pipeline}, we first introduce two graph-based classifiers to construct an iteratively refined sentence-entity graph (IRSE-Graph).
It is a weighted version of SE-Graph,
where pairwise ordering inforamtion is iteratively incorporated to update ss-edge weights.
Then, we adapt the conventional GRN to establish a neural sentence ordering model based on the final IRSE-Graph.

\subsection{The Definition of IRSE-Graph}
As an extension of SE-Graph, 
IRSE-Graph can be denoted as $G$$=$$(\boldsymbol{V}$$,$$\boldsymbol{E}$$,$$\boldsymbol{W})$, where $\boldsymbol{V}$ and $\boldsymbol{E}$ share the same definitions with those of SE-Graph. Particularly, in IRSE-Graph, each ss-edge $e_{i,i'}$ is a directed one with a weight $w_{i,i'}$$\in$$\boldsymbol{W}$ indicating the probability of sentence $s_i$ occurring before sentence $s_{i'}$.
Meanwhile, there must exist a corresponding ss-edge $e_{i',i}$ with the weight $w_{i',i}$$=$$1$$-$$w_{i,i'}$ denoting the probability of $s_{i}$ appearing after $s_{i'}$. For example, in Figure \ref{pipeline}, for two linked sentence nodes $v_1$ and $v_2$,  there exist two ss-edges $e_{1,2}$ and $e_{2,1}$ with weights $w_{1,2}$ and $w_{2,1}$ respectively, both of which are iteratively updated during constructing IRSE-Graph.

\subsection{Constructing IRSE-Graph}\label{subsection:two classifier}
Inspired by \citet{Gui:EMNLP20},	we successively introduce two classifiers --- \emph{initial classifier} and \emph{iterative classifier} to construct IRSE-Graph. Both classifiers are constructed using slightly adapted GRN and utilized to deal with different scenarios, respectively.
In this way, we can fully exploit the potential of iterative classifier to predict better pairwise orderings. 
We will give a detail introduction to the slightly adapted GRN in Section \S\ref{subsection:IRSE-GRN encoding}.

To better understand the procedure of constructing IRSE-Graph, we provide the details in Algorithm \ref{code:iteration}. During this procedure, pairwise orderings are iteratively predicted and gradually incorporated to refine IRSE-Graph. Here we introduce a set \emph{VP}$^{(k)}$ to collect sentence node pairs with uncertain pairwise orderings at the $k$-th iteration.

First, we bulid an initial classifier based on the initial IRSE-Graph, where the learned sentence representations are used to predict pairwise orderings between any two linked sentences only once (\textbf{Lines 2-6}).
Note that in the initial IRSE-Graph, all weights of ss-edges are set to 0.5. In this case, IRSE-Graph degrades to the conventional SE-Graph. 
Concretely,
for any two linked sentence nodes $v_i$ and $v_{i'}$, we concatenate their vector representations $\boldsymbol{\kappa}_{i}$ and $\boldsymbol{\kappa}_{i'}$ as $\left[\boldsymbol{\kappa}_{i};\boldsymbol{\kappa}_{i'}\right]$ and $\left[\boldsymbol{\kappa}_{i'};\boldsymbol{\kappa}_{i}\right]$,
which are fed into an MLP classifier to obtain two probabilities. Then, we normalize and convert these two probabilities into ss-edge weights $w_{i,i'}$ and $w_{i',i}$.
If both $w_{i,i'}$ and $w_{i',i}$ are within a prefixed interval $\left[\delta_{min}, \delta_{max}\right]$, we consider $(v_i,v_{i'})$ as a sentence node pair with uncertain pairwise ordering and add it into \emph{VP}$^{(0)}$.
Moreover, we replace both $w_{i,i'}$ and $w_{i',i}$ with 0.5, indicating that they will be repredicted in the next iteration.

\begin{algorithm}[!t] \small 
	\begin{algorithmic}[1]
		\Require ~~the initial IRSE-Graph: $G$$=$$(\boldsymbol{V},\boldsymbol{E},\boldsymbol{W})$ with all $w_{i,i'}$$=$$0$; two thresholds: $\delta_{min},\delta_{max}$

		\Ensure the final IRSE-Graph: $G=(\boldsymbol{V},\boldsymbol{E},\boldsymbol{W})$
		\State\label{code:1} \emph{VP}$^{(0)}$ $\gets\emptyset$
		\State $\{\boldsymbol{\kappa}_i\}_{i=1}^{I} \gets$ GRN$(G)$
		\For{any linked sentence node pair $(v_i, v_{i'})$ \textbf{\&\&} $i$$<$$i'$}
		\State $w_{i,i'} \gets$ InitialClassifer$([\boldsymbol{\kappa}_i;\boldsymbol{\kappa}_{i'}])$
		\State $w_{i',i} \gets$ InitialClassifer$([\boldsymbol{\kappa}_{i'};\boldsymbol{\kappa}_i])$
		\State $w_{i,i'},w_{i',i} \gets$ Normalize$(w_{i,i'},w_{i',i})$
		\If{$ \delta_{min} \le w_{i,i'} \le \delta_{max}$ }
		\State \emph{VP}$^{(0)}$$\gets$ \emph{VP}$^{(0)}$$\cup \{(v_i,v_{i'})\}$
		\State $w_{i,i'} \gets 0.5$, $w_{i',i} \gets 0.5$
		
		\EndIf
		\EndFor
		\State $k \gets 0$
		\Repeat
		\State \emph{VP}$^{(k+1)}$$\gets \emptyset$ 
		\State $\{\boldsymbol{\kappa}_i\}_{i=1}^I \gets$ GRN$(G)$ \label{code:15}
		\For{$(v_i,v_{i'}) \in$ \emph{VP}$^{(k)}$}
		\State $w_{i,i'} \gets$ IterativeClassifer$([\boldsymbol{\kappa}_i;\boldsymbol{\kappa}_{i'}])$
		\State $w_{i',i} \gets$ IterativeClassifer$([\boldsymbol{\kappa}_{i'};\boldsymbol{\kappa}_i])$
		\State $w_{i,i'},w_{i',i} \gets$ Normalize$(w_{i,i'},w_{i',i})$	
		\If{$ \delta_{min} \le w_{i,i'} \le \delta_{max}$ }
		\State \emph{VP}$^{(k+1)}$$\gets$\emph{VP}$^{(k+1)}$ $\cup \{(v_i,v_{i'})\}$	
		\State $w_{i,i'} \gets 0.5$, $w_{i',i} \gets 0.5$ 
		\EndIf
		\EndFor
		\State $k \gets k+1$        
		\Until{\emph{VP}$^{(k+1)}==$\emph{VP}$^{(k)}$ \textbf{||} \emph{VP}$^{(k)}==\emptyset$} \\
		\Return $G$
	\end{algorithmic}
	\caption{\label{code:iteration} The procedure of constructing \\ IRSE-Graph}
\end{algorithm}

In the following, we also construct an iterative classifier based on IRSE-Graph.
However, in an easy-to-hard manner, we use iterative classifier to perform pairwise ordering predictions,
where the ss-edge weights of IRSE-Graph are continously updated with previously-predicted pairwise orderings with high confidence (\textbf{Lines 13-26}).
By doing so, graph representations can be continously refined for better subsequent predictions.
More specifically, the $k$-th iteration of this classifier mainly involve three steps:

In \textbf{Step 1},
based on the current IRSE-Graph,
we employ the adapted GRN to conduct graph encoding to learn sentence representations (\textbf{Line 15}).

In \textbf{Step 2},
on the top of learned sentence representations, we stack an MLP classifier to predict pairwise orderings for sentence node pairs in \emph{VP}$^{(k)}$ (\textbf{Lines 16-19}).
Likewise, we collect sentence node pairs with uncertain pairwise orderings to form \emph{VP}$^{(k+1)}$,
and reassign their corresponding ss-edge weights as 0.5, so as to avoid the negative effect of these uncertain ss-edge weights during the next iteration (\textbf{Lines 20-23}).

In \textbf{Step 3}, if \emph{VP}$^{(k+1)}$ is equal to \emph{VP}$^{(k)}$ or empty, we believe the learning of IRSE-Graph $G$ has converged and thus return it (\textbf{Lines 26-27}).

\begin{figure}[t]
	\centering
	\includegraphics[width=1\linewidth]{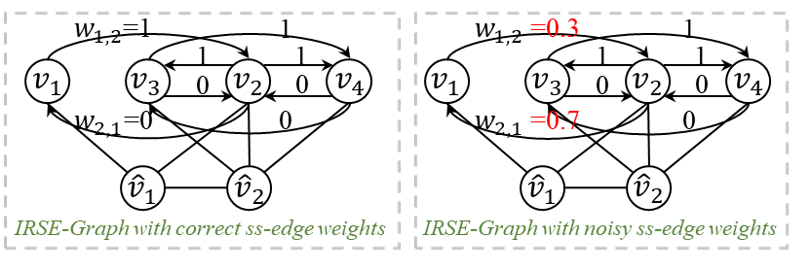}
	\caption{ \label{fig:replace} Introducing noisy ss-edge weights into IRSE-Graph.}

\end{figure}

Although both of our classifiers are constructed using IRSE-Graph, their training procedures are slightly different.
As for initial classifier, we directly train it on the initial IRSE-Graph without any pairwise ordering information (all ss-edge weights are set to 0.5).
By contrast, we train iterative classifier on IRSE-Graph with partial pairwise orderings. To enable iterative classifier generalizable to any IRSE-Graph with partial predicted pairwise orderings, 
we first set all ss-edge weights to 1 or 0 according to their ground-truth pairwise orderings,
and then train the classifier to correctly predict pariwise orderings for other pairs. 
Concretely, if $\boldsymbol{s}_i$ appears before $\boldsymbol{s}_{i'}$, we set $w_{i,i'}$$=$$1$ and $w_{i',i}$$=$$0$, vice versa.
For example, in the left part of Figure \ref{fig:replace}, the ground-truth sentence sequence is $s_1$,$s_2$,$s_3$,$s_4$, and thus we assign the ss-edge weights of linked sentence node pairs $(v_1,v_2)$, $(v_3,v_2)$, $(v_3,v_4)$, $(v_2,v_4)$ as follows:
$w_{1,2}$$=$$1$, $w_{2,3}$$=$$1$, $w_{2,4}$$=$$1$, $w_{3,4}$$=$$1$, and $w_{2,1}$$=$$0$, $w_{3,2}$$=$$0$, $w_{4,2}$$=$$0$, $w_{4,3}$$=$$0$.

Moreover, to enhance the robustness of the iterative classifier, we randomly select a certain ratio $\eta$ of sentence pairs and assign their ss-edges with incorrect weights.
Let us revisit Figure \ref{fig:replace}, for the randomly selected sentence node pair $(v_1,v_2)$, we assign ss-edges weights $w_{1,2}$ and $w_{2,1}$ with randomly generated noisy values 0.3 and 0.7 respectively.
In this way, we expect that iterative classifier can conduct correct predictions even given incorrect previously-predicted pairwise orderings.

\subsection{IRSE-Graph Sentence Ordering Model}\label{subsection:IRSE-GRN encoding}
Finally,
following the conventional SE-GRN \cite{Yin:IJCAI19,YinLSP_JAIR20}, we construct a graph-based sentence ordering model. Note that 
the above two classifiers and our sentence ordering model are all based on IRSE-Graph rather than the conventional SE-Graph, 
which makes the standard GRN unable to be applied directly. To deal with this issue,
we slightly adapt GRN to utilize pairwise ordering information for graph encoding.
Specifically, we adapt Equation \ref{eq:egraph_s_mess} to incorporate ss-edge weights into the message aggregation of sentence-level nodes: 

\begin{small}
	\vspace{-10pt}
	\begin{equation} \label{eq:irseraph_s_mess}
		\begin{aligned}
			\boldsymbol{m}_i^\text{($l$)}=\sum_{v_{i^\prime} \in N_i}&w_{i,i'}\cdot w(\boldsymbol{\kappa}_{i}^\text{($l$-1)},\boldsymbol{\kappa}_{i^\prime}^\text{($l$-1)})\boldsymbol{\kappa}_{i^\prime}^{\text{($l$-1)}}, \\
			w(\boldsymbol{\kappa}_{i}^\text{($l$-1)},\boldsymbol{\kappa}_{i^\prime}^\text{($l$-1)})&=
			\sigma(W_g[\boldsymbol{\kappa}_{i}^\text{($l$-1)};\boldsymbol{\kappa}_{i^\prime}^\text{($l$-1)}]).
		\end{aligned}
	\end{equation}
\end{small}Here $\sigma$ denotes sigmoid function and $W_g$ is learnable parameter matrix. Equation \ref{eq:irseraph_s_mess} expresses that the sentence-level aggregation should consider not only the semantic representations of the two involved sentences, but also the relative ordering between them.  In addition, other Equations are the same as those of conventional GRN, which have been described in Section \S\ref{subsection grn encoding}.

\section{Experiment}

\subsection{Setup}

\paragraph{Datasets.}

Following previous work \cite{Yin:AAAI20,Cui:emnlp18,YinLSP_JAIR20}, we carry out experiments on five benchmark datasets:

\begin{itemize}
	\vspace{-5pt}
	\setlength{\itemsep}{2pt}
	\setlength{\parskip}{0pt}
	\item \textbf{SIND, ROCStory}. SIND \cite{huang-EtAl:2016:N16-1} is a visual storytelling dataset and ROCStory \cite{mostafazadeh-etal-2016-corpus} is about commonsense stories. Both two datasets are composed of 5-sentence stories and randomly split by 8:1:1 for the training/validation/test sets. 
	
	\item \textbf{NIPS Abstract, AAN Abstract, arXiv Abstract}. These three datasets consist of abstracts from research papers, which are collected from NIPS, ACL anthology and arXiv, respectively \cite{Radev,Chen16}.
	The partitions for training/validation/test of each dataset are as follows: NIPS Abstract: 2,427/408/377, AAN Abstract: 8,569/962/2,626, arXiv Abstract: 884,912/110,614/110,615 for the training/validation/test sets.
	\vspace{-5pt}
\end{itemize}

\paragraph{Settings.}
For fair comparison, we use the same settings as our most related baseline SE-GRN \cite{YinLSP_JAIR20} for our model and its variants. Specifically, we apply 100-dimensional GloVe  word embeddings, and set the sizes of Bi-LSTM hidden states, sentence node states, and entity node states as 512, 512 and 150, respectively. The recurrent step of GRN is 3. We empirically set thresholds $\delta_{min}$ and $\delta_{max}$ as 0.2 and 0.8, and set $\eta$ as 20\%, 15\%, 25\%, 15\%, 15\% according to accuracies of initial classifier on validation sets. Besides, we individually set the coefficient $\lambda$ (See Equation 18 in \cite{Yin:AAAI20}) as 0.5, 0.5, 0.2, 0.4, 0.5 on the five datasets. We adopt Adadelta \cite{ADADELTA} with $\epsilon$ = $10^{-6}$, $\rho$ = $0.95$ and initial learning rate 1.0 as the optimizer. We employ L2 weight decay with coefficient $10^{-5}$, batch size of 16 and dropout rate of 0.5. 

When constructing our model based on BERT, we use the same settings as \cite{Cui:emnlp20}. Concretely, we set sizes of hidden states and node states to 768, the learning rate of BERT as 3e-3, the batch size as 16, 32, 128, 128, 64 for the five datasets.

\begin{table*}[]\scriptsize
	\centering
	\setlength{\tabcolsep}{1mm}{
		\begin{tabular}{lccc|ccc|ccc|ccc|ccc}  
			\toprule
			Model &\multicolumn{3}{c}{NIPS Abstract} &\multicolumn{3}{c}{AAN Abstract} &\multicolumn{3}{c}{SIND} & \multicolumn{3}{c}{ROCStory} &\multicolumn{3}{c}{arXiv Abstract}\\
			\cline{1-16}
			& Acc & $\tau$ & PMR & Acc & $\tau$ & PMR & Acc & $\tau$ & PMR & Acc & $\tau$ & PMR & Acc & $\tau$ & PMR \\
			\hline
			Pairwise Model \cite{Chen16}$^\dagger$		&-&-&- &-&-&- &-&-&- &-&-&- &- & 66.00 & 33.43 \\
			LSTM+PtrNet \cite{Gong16}$^\dagger$      & 50.87 & 67.00 &- & 58.20 & 69.00 &- & -     & 48.42 & 12.34 & - & -     & -     & - & 71.58 & 40.44 \\
			V-LSTM+PtrNet \cite{Logeswaran:aaai18}$^\dagger$	& 51.55 & 72.00 &- & 58.07 & 73.00 &- &-&-&- &-&-&- &- & - & -\\
			ATTOrderNet \cite{Cui:emnlp18}$^\dagger$		& 56.09 & 72.00 &- & 63.24 & 73.00 &- & -     & 49.00 & 14.01 & - & -     & -     & - & 73.00 & 42.19 \\
			HAN \cite{Wang:AAAI2019}$^\dagger$				&-&-&- &-&-&- &-& 50.00 & 15.01 &-& 73.00 &39.62 &- & 75.00 & 44.55 \\
			SE-GRN \cite{Yin:IJCAI19}$^\dagger$ 		& 57.27 & 75.00 &- & 64.64 & 78.00 &- 	& -     & 52.00 & 16.22 & - & -     & -     & - & 75.00 & 44.33 \\
			SE-GRN \cite{YinLSP_JAIR20}		& 58.25 & 76.49 &25.73 & 65.06 & 78.60 & 44.87	& 49.58     & 53.16 & 17.17 & 68.96 & 75.46  & 42.67     & 59.07 & 75.74 & 44.72 \\
			ATTOrderNet+FHDecoder \cite{Yin:AAAI20}$^\dagger$     &-&-&- &-&-&-   & -     & 53.19 & 17.37 & - & 76.81 & 46.00 & - & 76.54 & 46.58\\
			TGCM \cite{Oh:emnlp19}$^\dagger$ &59.43&75.00&31.44 &65.16&75.00&36.69 &38.71&15.18&53.00 &-&-&- &58.31&75.00&44.28 \\
			\hdashline[2pt/2pt]	
			
			RankTxNet \cite{Kumar:aaai20}$^\dagger$ &-&75.00&24.13 &-&77.00&39.18 &-&57.00&15.48 &-&76.00&38.02 &-&77.00&43.44 \\
			B-TSort \cite{Prabhumoye:ACL20}$^\dagger$ &61.48&81.00&32.59 &69.22&83.00&50.76 &52.23&60.00&20.32 &-&-&- &-&-&- \\
			ConsGraph \cite{Zhu:AAAI2021}$^\dagger$ &-&80.29&32.84 &-&82.36&49.81 & - & 58.56 & 19.07 & - & 81.22 & 49.52 & - & - & - \\
			
			BERSON \cite{Cui:emnlp20}$^\dagger$ &{73.87}&{85.00}&{48.01} &{78.03}&{85.00}&{59.79} & {58.91} & {65.00} & \bf{31.69} & {82.86} & {88.00} & {68.23} & {75.08} & {83.00} & {56.06} \\ 
			
			\hline
			IRSE-GRN         &63.14&80.45&32.63 &68.51&82.09&49.56	& 51.01 & 54.97 & 18.77 & 71.28 & 77.43 & 46.38 & 70.15 & 84.22 & 56.85 \\
			IRSE-GRN+FHDecoder    &73.62&87.45&50.19 &77.34&87.87&62.24	& 54.98 & 61.87 & 22.77 & 77.70 & 84.20 & 57.11 & 74.45 & 88.57 & 60.30 \\
			\hdashline[2pt/2pt]	
			IRSE-GRN+BERT+FHDecoder  &\bf{78.00}&\bf{90.35}&\bf{58.81} &\bf{82.07}&\bf{91.11}&\bf{68.93}	& \bf{59.08} & \bf{66.14} & 28.79 & \bf{83.77} & \bf{89.09} & \bf{69.06} & \bf{78.64} & \bf{90.30} & \bf{66.59} \\

			\bottomrule
			
	\end{tabular}}
	\vspace{-5pt}
	\caption{Main results on the sentence ordering task, where $\dagger$ indicates previously reported scores. Please note that RankTxNet, B-TSort and ConsGraph are pairwise models based on BERT, and the previous SOTA BERSON is also based on BERT and equipped with FHDecoder.}
	\label{tab:result}

\end{table*}

\paragraph{Baselines.}

To demonstrate the effectiveness of our model (IRSE-GRN), we compare it with SE-GRN \cite{YinLSP_JAIR20}. Besides, we report the performance of following sentence ordering models: 1) \textbf{Pairwise models}: 
Pairwise Model \cite{Chen16}, RankTxNet \cite{Kumar:aaai20}, and 
B-TSort \cite{Prabhumoye:ACL20}, ConsGraph \cite{Zhu:AAAI2021};
2) \textbf{Set-to-sequence models}: 
HAN \cite{Wang:AAAI2019}, 
LSTM+PtrNet \cite{Gong16}, 
V-LSTM+PtrNet \cite{Logeswaran:aaai18}, 
ATTOrderNet \cite{Cui:emnlp18}, 
TGCM \cite{Oh:emnlp19},
SE-GRN \cite{Yin:IJCAI19}, 
SE-GRN \cite{YinLSP_JAIR20}, 
ATTOrderNet+FHDecoder \cite{Yin:AAAI20} 
and BERSON \cite{Cui:emnlp20}.

Furthermore, to examine the compatibility of other technologies with our model, we report the performance of IRSE-GRN equipped with some effective components: 1) \textbf{IRSE-GRN+FHDecoder}. In this variant, we equip our model with FHDecoder \cite{Yin:AAAI20}, where pairwise ordering information is incorporated; 2) \textbf{IRSE-GRN+BERT+FHDecoder}. In addition to FHDecoder, we construct the sentence encoder based on BERT, where the mean-pooling outputs of all learned word representations are used to initialize sentence nodes.

\vspace{-5pt}
\paragraph{Evaluation Metrics.} 
Following previous work \cite{Oh:emnlp19,Cui:emnlp20,Prabhumoye:ACL20,Zhu:AAAI2021,YinLSP_JAIR20}, we use the following three metrics: 
1) \textbf{Kendall's Tau ($\tau$)}: Formally, this metric is calculated as 1- 2$\times$(\emph{number of inversions})/$\binom{I}{2}$, where $I$ denotes the sequence length and \emph{number of inversions} is the number of pairs in the predicted sequence with incorrect relative order \cite{Lapata:acl03};
2) \textbf{Perfect Match Ratio (PMR)}: This metric calculates the ratio of samples where the entire sequence is correctly predicted \cite{Chen16};
3) \textbf{Accuracy (Acc)}: This metric measures the percentage of sentences, whose absolute positions are correctly predicted \cite{Logeswaran:aaai18}.
\vspace{-5pt}
\subsection{Pairwise Ordering}

\begin{table}[t] \scriptsize
	\centering
	\begin{tabular}{lccc}  
		\toprule
		{Dataset}
		&\multicolumn{1}{l}{Initial Classifier} &\multicolumn{1}{l}{Initial + Iterative Classifiers} \\
		\midrule
		NIPS Abstract     & 80.46\% & \bf{86.32\%}   \\
		AAN Abstract	&84.53\% & \bf{86.74\%} \\
		SIND	& 77.72\% & \bf{83.55\%}  \\ 
		ROCstory 	& 87.59\% & \bf{92.23\%}  \\
		arXiv Abstract	& 84.09\% & \bf{86.82\%}  \\
		\bottomrule
	\end{tabular}
	\caption{The accuracies of our two classifiers on five test datasets.}
	\label{tab:pairwise}

\end{table}
Since pairwise ordering plays a crucial role in our proposed framework, we first compare the performance of different classifiers on various datasets. Table \ref{tab:pairwise} shows the experimental results. Obviously, the utilization of iterative classifier further benefits the predictions of pairwise orderings.

\subsection{Main Results}

Table \ref{tab:result} reports the overall experimental results of sentence ordering.
When incorporating BERT and FHDecoder into IRSE-GRN,
our model achieves SOTA performance on most of datasets.
Besides,
we arrive at the following conclusions:

\textbf{First}, IRSE-GRN significantly surpasses SE-GRN on all datasets (bootstrapping test, $p$$<$$0.01$), 
indicating that iteratively refining graph representations indeed benefit the ordering of input sentences. 

\textbf{Second}, IRSE-GRN+FHDecoder exhibits better performance than IRSE-GRN and all non-BERT baselines, which are shown above the upper dotted line of Table \ref{tab:result}, across datasets in different domains.
Therefore, we confirm that our framework is orthogonal to the current approach exploiting pairwise ordering information for decoder.

\textbf{Third}, when constructing our model based on BERT, IRSE-GRN+BERT+FHDecoder also outperforms all BERT-based baselines, such as ConsGraph, BERSON, achieving SOTA performance.
It can be known that our proposed framework is also effective when combining with pretrained language model.

\textbf{Finally}, we note that IRSE-GRN+BERT+FH-\\Decoder gains relatively marginal improvement on SIND and ROCStory, and performs worse than BERSON in PMR on SIND. We speculate that there exist less ss-edges on these two datasets, resulting in that our proposed framework can not achieve its full potential. Specifically, average edge numbers of SIND and ROCStory are 2.85 and 5.66 respectively, far fewer than 16.60, 10.86 and 16.73 on NIPS Abstract, ANN Abstract and arXiv Abstract.

\begin{figure}
	\centering
	\includegraphics[width=1\linewidth]{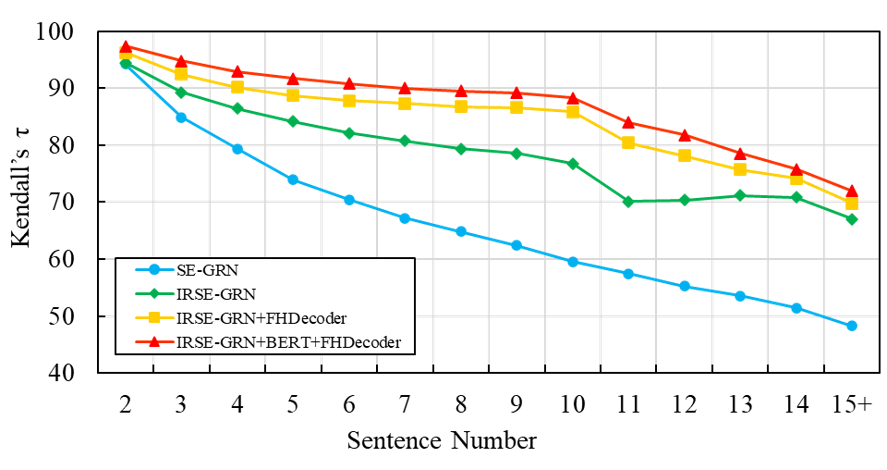}
	\caption{ \label{fig:SentenceNumber} The Kendall's $\tau$ of different models with respect to different sentence numbers on the arXiv abstract test set.}

\end{figure}
%
%
%

Besides, since it is a challenge to order longer paragraphs, we investigate the Kendall's $\tau$ of our models and SE-GRN with respect to different sentence numbers, as shown in Figure \ref{fig:SentenceNumber}. Overall, all models degrade with the increase of sentence number. However, our model and its two enhanced versions always exhibit better performance than SE-GRN.

\subsection{Predictions of the First and Last Sentences}
As mentioned in previous studies \cite{Gong16,Chen16,Cui:emnlp18,Oh:emnlp19}, the first and last sentences are very important in a paragraph. Following these studies, we compare models by conducting experiments to predict the first and last sentences.

As displayed in Table \ref{tab:headtail}, IRSE-GRN surpasses all non-BERT baselines, and IRSE-GRN+BERT+ FHDecoder wins against BERTSON. These results are consistent with those reported in Table \ref{tab:result}, further demonstrating the effectiveness of our model.

\subsection{Ablation Study}
We conduct several experiments to investigate the impacts of our proposed components on ROCstory dataset and arXiv dataset which are the two largest datasets. All results are provided in Table \ref{ablation}, where we draw the following conclusions:

\textbf{First}, using only iterative classifier, IRSE-GRN(w/o initial classifier) performs worse than IRSE-GRN.
This result proves that iterative classifier fails to predict well from scratch and the pairwise ordering predicted by initial classifier is beneficial to construct a well-formed graph representation for iterative classifier. 

\textbf{Second}, when the iteration number \emph{k} is set as 1, the performance of IRSE-GRN decreases.
Moreover, if we remove iterative classifier, the performance of IRSE-GRN becomes even worse.
Therefore, we confirm that the iterative predictions of pairwise ordering indeed benefit the learning of graph representations.

\textbf{Finally}, the result in the last line indicates that removing noisy weights leads to a significant performance
drop. It suggests that the utilization of noisy weights is useful for the training of iterative classifier, which makes our model more robust.

\begin{table}[t] \scriptsize
	\centering
	\setlength{\tabcolsep}{1mm}{
		
		\begin{tabular}{lcccccc}  
			\toprule
			{Model} 	&\multicolumn{2}{c}{SIND} &\multicolumn{2}{c}{arXiv Abstract} \\
			\cmidrule{1-5}
			& head & tail & head & tail  \\
			\midrule
			Pairwise Model \cite{Chen16}$^\dagger$ & -& - & 84.85 & 62.37 \\
			LSTM+PtrNet \cite{Gong16}$^\dagger$ & 74.66 & 53.30 & 90.47 & 66.49 \\
			ATTOrderNet \cite{Cui:emnlp18}$^\dagger$ & 76.00 & 54.42 & 91.00 & 68.08 \\
			SE-GRN \cite{Yin:IJCAI19}$^\dagger$	  & 78.12 & 56.68 & 92.28 & 70.45 \\
			SE-GRN \cite{YinLSP_JAIR20}  & 79.01 & 57.27 & 92.23 & 70.46 \\
			ATTOrderNet+FHDecoder \cite{Yin:AAAI20}$^\dagger$ & 78.08 & 57.32 & 92.76 & 71.49 \\
			TGCM \cite{Oh:emnlp19}$^\dagger$	  & 78.98 & 56.24 & 92.46 & 69.45 \\
			\hdashline[2pt/2pt]
			RankTxNet \cite{Kumar:aaai20}$^\dagger$	  & 80.32 & 59.68 & 92.97 & 69.13 \\
			B-Tsort \cite{Prabhumoye:ACL20}$^\dagger$	& 78.06 & 58.36 & - & - \\ 
			ConsGraph \cite{Zhu:AAAI2021}$^\dagger$	   & 79.80 & 60.44 & - & - \\
			BERSON	\cite{Cui:emnlp20}$^\dagger$	  & 84.95 & 64.87 & 94.75 & 76.69 \\
			\midrule
			IRSE-GRN 		  & 78.62 & 59.11 & 94.46 & 80.97 \\
			IRSE-GRN+FHDecoder  & 82.87 & 64.15 & 96.09 & 85.04 \\
			\hdashline[2pt/2pt]
			IRSE-GRN+BERT+FHDecoder   & \bf{86.21} & \bf{67.14} & \bf{98.23} & \bf{88.33} \\
			\bottomrule
	\end{tabular}}
	
	\caption{The ratios of correctly predicting first and last sentences on arXiv Abstract and SIND. $\dagger$ indicates previously reported scores.}
	\label{tab:headtail}

\end{table}

\begin{table}[t] 
	\centering \scriptsize
	\setlength{\tabcolsep}{0.8mm}{
		\begin{tabular}{lc|cc|c|cc}  
			\toprule
			{Model}
			& \multicolumn{3}{c}{ROCStory} &\multicolumn{3}{c}{arXiv Abstract}\\
			\cline{1-7}
			&Pairwise  & $\tau$  & PMR & Pairwise & $\tau$ & PMR \\
			\hline
			IRSE-GRN	 						& \bf{92.23} & \bf{77.43} & \bf{46.38} & \bf{86.82} & \bf{84.22} & \bf{56.85} \\
			\qquad w/o initial classifier  & 88.96 & 77.31 & 45.06 & 77.90 & 81.03 & 51.65 \\
			\qquad iterative number $k$=1	             & 91.73 & 77.21 & 46.13 & 86.22 &  83.89  & 56.03  \\
			\qquad w/o iterative classifier 	 & 87.59 & 75.98 & 44.14 & 84.09 & 83.24 & 55.05 \\
			\qquad w/o noise	             & 90.42 & 77.06 & 46.02 & 80.45 &  82.23  & 53.10  \\
			\bottomrule
		\end{tabular}
		\caption{\label{ablation} Ablation study on the impacts of our proposed components on ROCStory dataset and arXiv abstract dataset.}}

\end{table}

\subsection{Summary Coherence Evaluation}

\begin{table}[t] 
	\centering \small
	\begin{tabular}{lcccccc}  
		\toprule
		{Model} 	&\multicolumn{2}{c}{Coherence}  \\
		\midrule
		SE-GRN \cite{YinLSP_JAIR20}  &46.71 & 59.47 \\
		\midrule
		IRSE-GRN 		 &47.48 &60.01  \\
		IRSE-GRN+FHDecoder    &49.84 & 61.81  \\
		IRSE-GRN+BERT+FHDecoder    &51.01  & 62.87 \\
		\bottomrule
	\end{tabular}
	\caption{Coherence probabilities of summaries reordered by different models using weights of 0.8 (left) and 0.5 (right). }
	\label{tab:summary}
\end{table}

\begin{table*}[bht] 
	\centering \small
	\setlength{\tabcolsep}{1.5mm}{
	\begin{tabular}{lcc|cc|cc|cc}  
		\toprule
		{Dataset}
		& \multicolumn{2}{c}{SE-GRN} &\multicolumn{2}{c}{IRSE-GRN} &\multicolumn{2}{c}{IRSE-GRN+FHDecoder} &\multicolumn{2}{c}{IRSE-GRN+BERT+FHDecoder}\\
		\cline{1-9}
		&Runtime  & \#Params   & Runtime & \#Params & Runtime & \#Params & Runtime & \#Params  \\
		\hline
		NIPS abstract	& 6s & 23.9M & 6.2s & 24.0M &18s&25.0M   &29s&128.0M \\
		AAN abstract  & 31s & 23.9M & 32.5s & 24.0M &1min8s& 25.0M    &1min20s&128.0M\\
		SIND	      & 1min6s & 23.9M  & 1min9s & 24.0M &2min3s&  25.0M     &2min16s&128.0M  \\
		ROCStory 	 & 2min & 23.9M & 2min5s & 24.0M &4min2s& 25.0M   &4min42s&128.0M \\
		arXiv abstract	& 25min & 23.9M & 27min57s & 24.0M &46min&  25.0M   &56min& 128.0M \\
		\bottomrule
	\end{tabular}}
	\caption{\label{tab:TimePara:base} The runtime on the validation sets and the numbers of parameters for our enhanced models and baseline.}
\end{table*}

Following previous studies \cite{Barzilay:ACL05,Nayeem:ACL17}, we further inspect the validity of our proposed framework via multi-document summarization. 
Concretely, we train different neural sentence ordering models on a large-scale summarization corpus \cite{Fabbri:ACL19}, and then individually use them to reorder the small-scale summarization data of DUC2004 (Task2).
Finally, we use coherence probability proposed by \cite{Nayeem:ACL17} to evaluate the coherence of summaries.
In this group of experiments, we conduct experiments using different weights: 0.5 and 0.8, as implemented in \cite{Nayeem:ACL17} and \cite{Yin:AAAI20} respectively.

The results are reported in Table \ref{tab:summary}. We can observe that the summaries reordered by IRSE-GRN and its variants achieve higher coherence probabilities than baseline, verifying the effectiveness of our proposed framework in the downstream task.

\subsection{Further Experiment Results}

To provide more experimental results, we summarize  the runtime on the validation sets and the numbers of parameters for our enhanced models and baseline SE-GRN in Table \ref{tab:TimePara:base}.

\section{Conclusion}

In this work, we propose a novel sentence ordering framework that  makes better use of pairwise orderings for graph-based sentence ordering. Specifically, we introduce two classifiers to iteratively predict pairwise orderings, which are gradually incorporated into the graph as edge weights. Then, based on this refined graph, we construct a graph-based sentence ordering model. Experiments on five datasets demonstrate not only the superiority of our model over baselines, but also the compatibility to other modules utilizing pairwise ordering information. Moreover, when equipped with BERT and FHDecoder, our enhanced model achieves SOTA performance across datasets. 

In the future, we plan to explore more effective GNN for sentence ordering.
In particular, we will improve our model by iteratively merging nodes to refine the graph representation.

\section*{Acknowledgment}
The project was supported by  
National Key Research and Development Program of China (No. 2020AAA0108004), 
National Natural Science Foundation of China (No. 61672440), 
Natural Science Foundation of Fujian Province of China (No. 2020J06001),
and Youth Innovation Fund of Xiamen (No. 3502Z20206059).
We also thank the reviewers for their insightful comments.

\bibliography{anthology,custom}
\bibliographystyle{acl_natbib}

\end{document}